# Analysis of Speedups in Parallel Evolutionary Algorithms for Combinatorial Optimization

Jörg Lässig[†] and Dirk Sudholt[‡]

[†] EAD Group, University of Applied Sciences Zittau/Görlitz, 02826 Görlitz, Germany
[‡] CERCIA, University of Birmingham, Birmingham B15 2TT, UK

**Abstract.** Evolutionary algorithms are popular heuristics for solving various combinatorial problems as they are easy to apply and often produce good results. Island models parallelize evolution by using different populations, called islands, which are connected by a graph structure as communication topology. Each island periodically communicates copies of good solutions to neighboring islands in a process called migration. We consider the speedup gained by island models in terms of the parallel running time for problems from combinatorial optimization: sorting (as maximization of sortedness), shortest paths, and Eulerian cycles. Different search operators are considered. The results show in which settings and up to what degree evolutionary algorithms can be parallelized efficiently. Along the way, we also investigate how island models deal with plateaus. In particular, we show that natural settings lead to exponential vs. logarithmic speedups, depending on the frequency of migration.

## 1 Introduction

Evolutionary algorithms (EAs) are popular heuristics for various combinatorial problems as they often perform better than problem-specific algorithms with proven performance guarantees. They are easy to apply, even in cases where the problem is not well understood or when there is not enough time or expertise to design a problem-specific algorithm. Another advantage is that EAs can be parallelized easily [17]. This is becoming more and more important, given the development in computer architecture and the rising number of CPU cores. Developing efficient parallel metaheuristics is a very active research area [1,16].

A simple way of using parallelization is to use so-called offspring populations: new solutions (offspring) are created and evaluated simultaneously on different processors. Island models use parallelization on a higher level. Subpopulations, called islands, which are connected by a graph structure, evolve independently for some time and periodically communicate copies of good solutions to neighboring islands in a process called migration. Migration is typically performed every $\tau$ iterations, the parameter $\tau$ being known as *migration interval*. A slow spread of information typically yields a larger diversity in the system, which can help for optimizing multimodal problems. For other problems a rapid spread of information (like setting $\tau = 1$ and migrating in every iteration) is beneficial, assuming low communication costs [9].

Despite wide-spread applications and a long history of parallel EAs, the theory of these algorithms is lagging far behind. Present theoretical work only concerns the study of the spread of information, or *takeover time*, in isolated and strongly simplified models (see, e.g., [13]) as well as investigations of island models on constructed test functions [7,8,10]. It is agreed that more fundamental research is needed to understand when and how parallel EAs are effective [15].

In this work we consider the speedup gained by parallelization in island models on illustrative problems from combinatorial optimization. The question is in how far using $\mu$ islands (each running an EA synchronously and in parallel) can decrease the number of iterations until a global optimum is found, compared to a single island. The number of iterations for such a parallel process is called *parallel optimization time* [9]. If the expected parallel optimization time is by a factor of $\Theta(\mu)$ smaller than the expected time for a single island, we speak of an (asymptotic) linear speedup. A linear speedup implies that a parallel and a sequential algorithm have the same total computational effort, but the parallel time for the former is smaller. We are particularly interested in the range of $\mu$ for which a linear speedup can be guaranteed. This degree of parallelizability depends on the problem, the EA running on the islands, and the parameters of the island model. Our investigation gives answers to the question how many islands should be used in order to achieve a reasonable speedup. Furthermore, it sheds light on the impact of design choices such as the communication topology and the migration interval $\tau$ as the speedup may depend heavily on these aspects.

Following previous research on non-parallel EAs [12], we consider various well-understood problems from combinatorial optimization: sorting as an optimization problem [14] (Section 4), the single-source shortest path problem [2,5] (Section 5), and the Eulerian cycle problem [3,4,11] (Section 6). As in previous studies, the purpose is not to design more efficient algorithms for well-known problems. Instead, the goal is to understand how general-purpose heuristics perform when being applied to a broad range of problems. The chosen problems contain problem features that are also present in more difficult, NP-hard problems. In particular, the Eulerian cycle problem contains so-called *plateaus*, that is, regions of the search space with equal objective function values. Our investigations pave the way for further studies that may include NP-hard problems.

For the sake of readability, some proofs are put in an appendix.

## 2   Preliminaries

Island models evolve separate subpopulations—islands—independently for some time. Every $\tau$ generations at the end of an iteration, or *generation* using common language of EAs, copies of selected search points or *individuals* are sent as migrants to neighbored islands. Depending on their objective value $f$, or *fitness*, migrants are in the target island's population after selection. The neighborhood of the islands is defined by a topology, a directed graph with the islands as nodes.

Algorithm 1 presents a general island model, formulated for maximization. Like in many previous studies for combinatorial optimization [12], we consider

**Algorithm 1** Island model

Let $t := 0$. For all $1 \leq i \leq \mu$ initialize population $P_0^i$ uniformly at random.
**repeat**
    For all $1 \leq i \leq \mu$ do in parallel
        Choose $x^i \in P_t^i$ uniformly at random.
        Create $y^i$ by mutation of $x^i$.
        Choose $z^i \in P_t^i$ with minimum fitness in $P_t^i$.
        **if** $f(y^i) \geq f(z^i)$ **then** $P_{t+1}^i = P_t^i \setminus \{z^i\} \cup \{y^i\}$ **else** $P_{t+1}^i = P_t^i$.
        **if** $t \bmod \tau = 0$ and $t > 0$ **then**
            Migrate copies of an individual with maximum fitness in $P_{t+1}^i$
                to all neighbored islands.
            Let $y^i$ be of maximum fitness among immigrants.
            Choose $z^i \in P_{t+1}^i$ with minimum fitness in $P_{t+1}^i$.
            **if** $f(y^i) > f(z^i)$ **then** $P_{t+1}^i = P_{t+1}^i \setminus \{z^i\} \cup \{y^i\}$.
    Let $t = t + 1$.

islands of population size only 1, running variants of the (1+1) EA or randomized local search (RLS). Both maintain a single search point and create a new search point in each generation by applying a mutation operator. This offspring replaces the current solution if its fitness is not worse. RLS uses local operators for mutation, while the (1+1) EA uses a stochastic neighborhood [12]. For the $\mu$-vertex complete topology $K_\mu$, the island model then basically equals what is known as (1+$\mu$) EA or (1+$\mu$) RLS, respectively, if we migrate in every generation ($\tau = 1$): the best of $\mu$ offspring competes with the parent as in the (1+1) EA.

We consider different topologies to account for different physical architectures and assume that the communication costs on the physical topology are so low that it allows us to focus on the parallel optimization time only. Unless specified otherwise, we assume $\tau = 1$.

## 3 Previous Work

The authors [9,10] presented general bounds for parallel EAs by generalizing the *fitness-level method* or *method of f-based partitions* (see Wegener [18]). The idea of the method is to divide the search space into sets $A_1, \ldots, A_m$ strictly ordered w.r.t. fitness: $A_1 <_f A_2 <_f \cdots <_f A_m$ where $A <_f B$ iff $f(a) < f(b)$ for every $a \in A, b \in B$. In addition, $A_m$ contains only global optima.

We say that a population-based algorithm (including populations of size 1) is in $A_i$ or on level $i$ if the current best individual in the population is in $A_i$. Elitist algorithms (defined as algorithms where the best solution in the population never worsens) can only increase the current level. The goal is to reach $A_m$. If $s_i$ is a lower bound on the probability of leaving $A_i$ towards any higher fitness level in one generation, the expected waiting time is at most $1/s_i$. As every level has to be left at most once, the expected optimization time is at most $\sum_{i=1}^{m-1} 1/s_i$.

The authors [9] generalized this method for island models that run elitist islands, for commonly used topologies. If migration is used in every generation, information about the current best fitness level is propagated to neighbored islands. This increases the number of islands searching for better fitness levels in parallel. The following theorem summarizes (a refinement of) our results.

**Theorem 1.** *Consider an island model with $\mu$ islands where each island runs an elitist EA. In every iteration each island sends copies of its best individual to all neighbored islands (i.e. $\tau = 1$). Each island incorporates the best out of its own individuals and its immigrants.*

*For every partition $A_1 <_f \cdots <_f A_m$ if $s_i$ is a lower bound for the probability that in one generation an island in $A_i$ finds a search point in $A_{i+1} \cup \cdots \cup A_m$ then the expected parallel optimization time is bounded by*

1. $2 \sum_{i=1}^{m-1} \frac{1}{s_i^{1/2}} + \frac{1}{\mu} \sum_{i=1}^{m-1} \frac{1}{s_i}$ *for every unidirectional ring (a ring with edges in one direction) or any other strongly connected topology,*
2. $3 \sum_{i=1}^{m-1} \frac{1}{s_i^{1/3}} + \frac{1}{\mu} \sum_{i=1}^{m-1} \frac{1}{s_i}$ *for every undirected grid or torus graph with side lengths at least $\sqrt{\mu} \times \sqrt{\mu}$,*
3. $m + \frac{1}{\mu} \sum_{i=1}^{m-1} \frac{1}{s_i}$ *for the complete topology $K_\mu$.*

Assuming the fitness-level bound for the time $\sum_{i=1}^{m-1} \frac{1}{s_i}$ of a single island is asymptotically tight, all three bounds yield an asymptotic linear speedup in case the first summands are each of at most the same order as the second summand.

Apart from the different constants 2 and 3, denser topologies yield better upper bounds than sparse ones. This makes sense as with the fitness level argumentation a rapid spread of information gives the best estimates for the time an improvement is found. The motivation for studying sparse topologies is that they have lower communication cost and they yield a larger diversity. An example where this diversity is beneficial will be given in Section 6.

## 4   Sorting

We start our investigations with the first combinatorial problem for which EAs have been analyzed. Scharnow, Tinnefeld, and Wegener [14] considered the classical sorting problem as an optimization problem: given a sequence of $n$ distinct elements from a totally ordered set, sorting is the problem of maximizing sortedness. W. l. o. g. the elements are $1, \ldots, n$, then the aim is to find the permutation $\pi_{\text{opt}}$ such that $(\pi_{\text{opt}}(1), \ldots, \pi_{\text{opt}}(n))$ is the sorted sequence.

The search space is the set of all permutations $\pi$ on $1, \ldots, n$. Two different operators are used for mutation. An exchange chooses two indices $i \neq j$ uniformly at random from $\{1, \ldots, n\}$ and exchanges the entries at positions $i$ and $j$. A jump chooses two indices in the same fashion. The entry at $i$ is put at position $j$ and all entries in between are shifted accordingly. For instance, a jump with $i = 2$ and $j = 5$ would turn $(1, 2, 3, 4, 5, 6)$ into $(1, 3, 4, 5, 2, 6)$.

The (1+1) EA draws $S$ according to a Poisson distribution with parameter $\lambda = 1$ and then performs $S + 1$ elementary operations. Each operation is either an exchange or a jump, where the decision is made independently and uniformly for each elementary operation. The resulting offspring replaces its parent if its

Analysis of Speedups in Parallel EAs for Combinatorial Optimization      5| Algorithm | INV | HAM, LAS, EXC |
|---|---|---|
| (1+1) EA | $O(n^2 \log n)$ [14] | $O(n^2 \log n)$ [14] |
| island model on ring | $O\left(n^2 + \frac{n^2 \log n}{\mu}\right)$ | $O\left(n^{3/2} + \frac{n^2 \log n}{\mu}\right)$ |
| island model on torus | $O\left(n^2 + \frac{n^2 \log n}{\mu}\right)$ | $O\left(n^{4/3} + \frac{n^2 \log n}{\mu}\right)$ |
| island model on $K_\mu/(1+\mu)$ EA | $O\left(n^2 + \frac{n^2 \log n}{\mu}\right)$ | $O\left(n + \frac{n^2 \log n}{\mu}\right)$ |

**Table 1.** Upper bounds for expected parallel optimization times for the (1+1) EA and the corresponding island model with $\mu$ islands for sorting $n$ objects.

fitness is not worse. The fitness function $f_{\pi_{\mathrm{opt}}}(\pi)$ describes the sortedness of $(\pi(1), \ldots, \pi(n))$. As in [14], we consider the following measures of sortedness:

- INV($\pi$) measures the number of pairs $(i, j), 1 \leq i < j \leq n$, such that $\pi(i) < \pi(j)$ (pairs in correct order)
- HAM($\pi$) measures the number of indices $i$ such that $\pi(i) = i$ (elements at the correct position),
- LAS($\pi$) equals the largest $k$ such that $\pi(i_1) < \cdots < \pi(i_k)$ for some $i_i < \cdots < i_k$ (length of the longest ascending subsequence),
- EXC($\pi$) equals the minimal number of exchanges (of pairs $\pi(i)$ and $\pi(j)$) to sort the sequence, leading to a minimization problem.

The expected optimization time of the (1+1) EA is $\Omega(n^2)$ and $O(n^2 \log n)$ for all fitness functions. The upper bound is tight for LAS, and it is believed to be tight for INV, HAM, and EXC as well [14]. Theorem 1 yields the following.

**Theorem 2.** *The expected parallel optimization times of the (1+1) EA and the corresponding island model with $\mu$ islands are as in Table 1.*

For INV, all topologies guarantee a linear speedup only in case $\mu = O(\log n)$ and the bound $O(n^2 \log n)$ for the (1+1) EA is tight. The other functions allow for linear speedups up to $\mu = O(n^{1/2} \log n)$ (ring), $\mu = O(n^{2/3} \log n)$ (torus), and $\mu = O(n \log n)$ ($K_\mu$), respectively (again assuming tightness, otherwise up to a factor of $\log n$). Note how the results improve with the density of the topology. HAM, LAS, and EXC yield much better guarantees for the island model than INV, though there is no visible performance difference for a single (1+1) EA.

## 5 Shortest Paths

We now consider parallel variants of the (1+1) EA for the single source shortest path problem (SSSP). Its complexity for the (1+1) EA has been first considered in [14]. An SSSP instance is given by an undirected connected graph with vertices $\{1, \ldots, n\}$ and a distance matrix $D = (d_{ij})_{1 \leq i, j \leq n}$ where $d_{ij} \in \mathbb{R}_0^+ \cup \{\infty\}$ defines the length value for given edges from node $i$ to node $j$. We are searching for shortest paths from a node $s$ (w.l.o.g. $s = n$) to each other node $1 \leq i \leq n-1$.

A candidate solution is represented as a *shortest paths tree*, a tree rooted at $s$ with directed shortest paths to all other vertices. We define a search point $x$ as vector of length $n-1$, where position $i$ describes the predecessor node $x_i$ of node $i$ in the shortest path tree. Note that infeasible solutions are possible in case the predecessors do not encode a tree. An elementary mutation chooses a vertex $i$ uniformly at random and replaces its predecessor $x_i$ by a vertex chosen uniformly at random from $\{1, \ldots, n\} \setminus \{i, x_i\}$. We call this a vertex-based mutation. The (1+1) EA creates an offspring using $S$ elementary mutations, where $S$ is chosen according to a Poisson distribution with $\lambda = 1$.

| Algorithm | vertex-based mutation [14] | edge-based mutation [5] |
|---|---|---|
| (1+1) EA | $\Theta(n^2\ell^*)$ [2] | $\Theta(m\ell^*)$ [5] |
| island model on ring | $O\left(n^{3/2}\ell^{1/2} + \frac{n^2\ell \ln(en/\ell)}{\mu}\right)$ | $O\left(m^{1/2}n^{1/2}\ell^{1/2} + \frac{m\ell \ln(en/\ell)}{\mu}\right)$ |
|  | $\longrightarrow \mu = O((n\ell)^{1/2})$ | $\longrightarrow \mu = O((m/n \cdot \ell)^{1/2})$ |
| island model on torus | $O\left(n^{4/3}\ell^{1/3} + \frac{n^2\ell \ln(en/\ell)}{\mu}\right)$ | $O\left(m^{1/3}n^{2/3}\ell^{1/3} + \frac{m\ell \ln(en/\ell)}{\mu}\right)$ |
|  | $\longrightarrow \mu = O((n\ell)^{2/3})$ | $\longrightarrow \mu = O((m/n \cdot \ell)^{2/3})$ |
| i. m. on $K_\mu$/(1+$\mu$) EA | $O\left(n + \frac{n^2\ell \ln(en/\ell)}{\mu}\right)$ | $O\left(n + \frac{m\ell \ln(en/\ell)}{\mu}\right)$ |
|  | $\longrightarrow \mu = O(n\ell)$ | $\longrightarrow \mu = O(m/n \cdot \ell)$ |

**Table 2.** Worst-case expected parallel optimization times for the (1+1) EA and the corresponding island model with $\mu$ islands for the SSSP on graphs with $n$ vertices and $m$ edges. The value $\ell$ is the maximum number of edges on any shortest path from the source to any vertex and $\ell^* := \max\{\ell, \ln n\}$. The second lines show a range of $\mu$-values yielding a linear speedup, apart from a factor $\ln(en/\ell)$.

The fitness function is defined as follows: Let $f(x) = (f_1(x), \ldots, f_{n-1}(x))$ and $f_i(x)$ code the length of the path from $s$ to $i$ if it is described by $x$ or $f_i(x) = \infty$ otherwise. The function $f(x)$ defines a partial order on the search points: $f(x) \leq f(x') \iff f_i(x) \leq f_i(x')$ for all $i \in \{1, 2, \ldots, n-1\}$. That defines a multi-objective minimization problem but there is exactly one Pareto optimal fitness vector. The multi-objective (1+1) EA chooses an initial search point $x$ uniformly at random and performs in each iteration a mutation step as described above. The new search point $x'$ is accepted if $f(x') \leq f(x)$.

The expected parallel optimization time can be bounded as follows. Partition the vertices into *layers* $1, \ldots, \ell$ such that the $j$-th layer contains all vertices having shortest paths of at most $j$ edges. When shortest paths have been found for all layers $1, \ldots, j$, shortest paths for vertices in layer $j + 1$ can be found by assigning the correct predecessor in a lucky mutation. The probability for making an improvement is at least $i/(en^2)$, in case $i$ vertices on layer $j$ still need to find the right predecessors [14]. Applying Theorem 1 to all layers and considering a worst-case for the arrangement of layers yields the following upper bounds.

**Theorem 3.** *The expected parallel optimization times of the multi-objective (1+1) EA and the corresponding island model with $\mu$ islands are bounded according to the first column of Table 2.*

The upper bounds for the island models with constant $\mu$ match the expected time of the (1+1) EA in case $\ell = O(1)$ or $\ell = \Omega(n)$ as then $\ell \ln(en/\ell) = \Theta(\ell^*)$. In other cases the upper bounds are off by a factor of $\ln(en/\ell)$. Table 2 also shows a range of $\mu$-value for which the speedup is linear (if $\ell = O(1)$ or $\ell = \Omega(n)$) or almost linear, that is, when disregarding the $\ln(en/\ell)$ term. Note how the possible speedups significantly increase with the density of the topology.

Doerr and Johannsen [5] presented the following novel mutation operator. Imagine predecessors to be represented by a set of edges such that for each vertex $v$ there is exactly one edge with end point $v$ in the set. Each elementary mutation consists of choosing an edge $(u,v)$ of the graph uniformly at random, adding it to the set, and removing the edge with end point $v$ from the set. This saves the (1+1) EA from assigning predecessors that are not connected to the vertex and it decreases the expected running time of the (1+1) EA by a factor of $O(m/n^2)$. By Lemma 3 in [5] the (lower bound for the) probability of making an improvement is increased to $i/(em)$. The resulting bounds for the (1+1) EA using this mutation operator are shown in the second column of Table 2.

Note that the ranges for possible speedups are never greater than for vertex-based mutations. This is because edge-based mutations are sometimes more efficient and never worse than vertex-based mutations in the (1+1) EA.

## 6 Eulerian Cycles

Given an undirected, loopless Eulerian graph, the task is to find an Eulerian cycle, that is, a graph traversal where each edge is traversed exactly once. A straightforward representation leads to plateaus, i. e., regions of equal fitness that have to be overcome by an EA. The performance of EAs on Eulerian cycles has been investigated in [3,4,6,11] where it has been shown that more sophisticated operators and representations lead to increasingly better performance.

Neumann [11] suggested a representation motivated by Hierholzer's algorithm. The idea of this algorithm is to subsequently concatenate cycles. This gives a *walk*, that is, a sequence of edges. When the walk includes all edges of the graph, an Eulerian cycle is created. Walks are represented by a permutation of the edges of the graph. The length of a walk $(e_1, e_2, \ldots, e_m)$ is the largest integer $\ell$ such that for all $1 \leq i \leq \ell - 1$ the edges $e_i$ and $e_{i+1}$ share a vertex. So, it is the length of a partial Euler walk. The first and last vertices of $e_1$ and $e_\ell$ are called *start* and *end* of the walk, resp. Neumann [11] as well as Doerr, Hebbinghaus, and Neumann [3] consider the length of the current walk as fitness and use jumps as mutation operators for RLS and the (1+1) EA. RLS always performs one jump, while the (1+1) EA chooses the number of jumps as in Section 4.

With the edge walk representation, fitness can be increased by appending a proper edge to the current walk. However, this operation is not always possible in case the current walk has closed a cycle. To see this, Neumann [11] defined the instance $G'$ as the concatenation of two cycles $C$ and $C'$, each consisting of $m/2$ edges, that share one common vertex $v^*$. This instance represents an asymptotic worst case for the time until an improvement is found.

If the current walk coincides with $C$, say, the current walk can only be extended by a single jump if it starts and ends with the vertex $v^*$. If it does not, the walk needs to be rotated until $v^*$ becomes start and end of the current walk. Rotations can be done by a jump with parameters $(1, m/2)$ or $(m/2, 1)$. As the fitness of all possible rotations of $C$ is equal, the algorithm has to search on a plateau. Since the two above jumps are equally likely, rotating $C$ with RLS corresponds to a fair random walk. With constant probability, the cycle needs to be rotated by a distance of $\Theta(m)$. This takes an expected number of $\Theta(m^2)$ steps of the random walk. As only two out of $m(m-1)$ possible jump operations are accepted, waiting for accepted jumps yields an additional factor of $\Theta(m^2)$. The expected optimization time of both RLS and (1+1) EA on $G'$ is $\Theta(m^4)$ [11].

| Mutation operator | RLS | par. RLS, frequent migr. | par. RLS, rare migr. |
|---|---|---|---|
| Unrestricted | $\Theta(m^4)$ [11,3] | $\Omega(m^4/(\log \mu))$ | $O(m^3 + 3^{-\mu} \cdot m^4)$ |
| Restricted, symmetric | $\Theta(m^3)$ | $\Omega(m^3/(\log \mu))$ | $O(m^2 + 3^{-\mu} \cdot m^3)$ |
| Restricted, asymmetric | $\Theta(m^2)$ | $O(m^2)$ | $O(m^2)$ |

**Table 3.** Expected parallel optimization times for RLS and the island model running RLS on $\mu = \text{poly}(m)$ islands with topology $T$ for computing an Eulerian cycle on $G'$. "Frequent migrations" is $\tau \cdot \text{diam}(T) \cdot \mu = O(m^2)$ for unrestricted jumps and $\tau \cdot \text{diam}(T) \cdot \mu = O(m)$ for symmetrically restricted ones, respectively. "Rare migrations" is $\tau \geq m^3$ and $\tau \geq m^2$, respectively.

$G'$ is a simple and natural instance as it represents the key features of the problem in a very clear way. It represents a worst-case for a single fitness level. It is not necessarily a global worst case as there is only one difficult fitness level, leaving a gap of $m$ to a general upper bound of $O(m^5)$ for all Eulerian graphs [11]. For simplicity, we focus on RLS instead of the (1+1) EA—here, both have equal asymptotic performance anyway [3,11]. Results are summarized in Table 3.

We give an example where parallelization does not reduce the parallel optimization time in any meaningful way. It can be shown that on $G'$ a single island with constant probability arrives at a solution where the current walk equals one of the two cycles and the cycle has to be rotated by a distance of $\Theta(m)$. If the migration interval is small enough (depending on the number of islands and the diameter of the topology), there is further a constant probability that this solution was spread throughout all islands. As only strictly better immigrants are considered for inclusion, all islands perform independent random walks. As the time for completing the random walk is highly concentrated, the expected time until the first island finds an improvement is still $\Omega(m^4/(\log \mu))$.

**Theorem 4.** *Consider the island model with an arbitrary strongly connected topology $T$ running RLS with jumps on each island. If $\tau \cdot \text{diam}(T) \cdot \mu = O(m^2)$ then the expected number of generations on $G'$ is at least $\Omega(m^4/(\log \mu))$.*

Using any polynomial number of islands only reduces the expected optimization time by at most a log-factor. However, in other settings parallelization can help dramatically. One positive effect of an island model is that islands can make different decisions on how to extend the current walk. On $G'$ this can make a difference between reaching the plateau and avoiding it completely.

In the beginning RLS typically evolves a walk on one of the two cycles $C$ and $C'$. If $v^*$, the vertex connecting $C$ and $C'$, is included in the current walk, the walk can either be extended towards the "opposite" cycle or it can move past $v^*$ and close the current cycle. In the former case a Eulerian cycle can be found easily by adding edges one-by-one. But in the latter case RLS has closed a cycle prematurely and it now has to rotate the walk to be able to include edges from the opposite cycle. This rotation dominates the expected running time.

Parallelization can help to make the right decision through independent evolution. If islands are run in parallel and if they evolve independently for at least $\tau \geq m^3$ generations, they tend to make independent decisions. This includes the case where no migration happens at all. The islands that have made the good decisions finish first, in expected time $\Theta(m^3)$. The remaining islands need $\Theta(m^4)$ steps in expectation. The probability of making a good decision is at least $2/3$ as a walk ending at $v^*$ can be extended by either of 3 edges, two of which lead to the opposite cycle; all 3 edges have the same probability for being added. Hence, the probability that a rotation—and time $\Theta(m^4)$—is needed is $3^{-\mu}$.

**Theorem 5.** *The island model running RLS on $\mu \leq \mathrm{poly}(m)$ islands, $\tau \geq m^3$, and an arbitrary topology optimizes $G'$ in expected $O(m^3 + 3^{-\mu} \cdot m^4)$ generations.*

The choice $\mu = \log_3 m$ leads to an expected parallel time of $O(m^3)$. This is a superlinear and, technically, even an exponential speedup. This is the first proof that island models can lead to a superlinear speedup on problems from combinatorial optimization.

The above result generalizes to instances where at $v^*$ more than two cycles come together. On other graphs the probability of not closing a cycle prematurely is exponentially small [3] and no speedups are possible. Details are omitted.

The results seen so far can be improved by restricting the mutation operator. The length of the current walk can only be increased in RLS if an edge jumps to either position 1 or $\ell + 1$. Choosing the second parameter uniformly from $\{1, \ell + 1\}$ (called a symmetric restriction) decreases all time bounds by a factor of $\Theta(m)$ (see Table 3). The authors of [3] introduced an asymmetric jump operator where the second parameter is fixed to 1, i.e., all edges are prepended to the current walk. This innocent-looking modification makes rotating cycles much easier as rotations are only possible in one direction. This removes the random-walk behavior, implying that the performance difference between frequent and rare migrations breaks down. It follows from Theorem 2 in [3] that then the island model running RLS with this operator finds an optimum on $G'$ in expected $O(m^2)$ generations, for any topology.

## 7  Conclusions

Considering speedups of island models has led to a surprising richness of results. For sorting linear speedups are possible, but the guarantees for parallelizability significantly depend on the measure of sortedness and the topology. The single-source shortest paths problem also allows for linear speedups, the maximum number of islands depending on the topology and the mutation operator. For Eulerian cycles results are inconclusive. Parallelization does not always help to speed up search on plateaus. However, it can help in some cases by avoiding plateaus if decisions where to extend the current edge walk are made correctly.

Also the parameters of the island model play a key role. On the same natural instance $G'$ speedups vary grossly from exponential up to $\mu = O(\log m)$ for rare or no migrations to at most logarithmic speedups, if migration is used too frequently and diversity is lost. Speedups also vary with the mutation operator.

**Acknowledgments:** The second author was supported by EPSRC grant EP/D052785/1.

## A   Appendix

The appendix contains all proofs that were omitted in the main part of the paper. In the following $T^{\mathrm{par}}$ denotes the parallel optimization time and $H(n)$ denotes the $n$-th harmonic sum.

### A.1   Proofs from Section 3

*Proof of Theorem 1.* For the third claim, observe that with $K_\mu$ all islands will be on the current best fitness level after migration. Then the probability of reaching a higher fitness level is at least $1 - (1 - s_i)^\mu$ and the expected time is bounded by

$$\frac{1}{1 - (1 - s_i)^\mu} \leq 1 + \frac{1}{\mu} \cdot \frac{1}{s_i}, \tag{1}$$

where the inequality was proposed by Jon Rowe (personal communication, 2011); it can be proven by a simple induction.

For the first bound, we claim that for every integer $k \leq \mu$ the expected time until fitness level $i$ is left is bounded by $k + \frac{1}{k} \cdot \frac{1}{s_i}$. The reason is that after $k - 1$ generations at least $k$ islands will be on the current best fitness level. This holds for the unidirectional ring and, in fact, for arbitrary strongly connected topologies. Along with (1), this proves the claim. Now, if $\mu \geq k := s_i^{-1/2}$ (ignoring rounding issues), the expected number of generations on fitness level $i$ is bounded by $k + 1/k \cdot s_i^{-1} = s^{-1/2} + s^{-1/2}$. If $\mu < s_i^{-1/2}$, we get for $k := \mu$ an upper bound of $\mu + 1/\mu \cdot s_i^{-1} > s_i^{-1/2} + 1/\mu \cdot s_i^{-1}$. Together, this proves the claimed bound.

Likewise, for the second bound after $2(\sqrt{k} - 1)$ iterations at least $k$ islands will be on the current fitness level as this time is sufficient to cover a rectangular area of $\sqrt{k} \times \sqrt{k}$ vertices in the topology. The expected time for leaving level $i$ is thus at most $2\sqrt{k} + \frac{1}{k} \cdot \frac{1}{s_i}$ for all $k \leq \mu$, again using (1). If $\mu \geq k := s_i^{-2/3}$, this gives $2s_i^{-1/3} + s_i^{-1/3} = 3s_i^{-1/3}$. Otherwise, $k := \mu$ yields a bound of $2\sqrt{\mu} + \frac{1}{\mu} \cdot \frac{1}{s_i} \leq 2s_i^{-1/3} + \frac{1}{\mu} \cdot \frac{1}{s_i}$. □

### A.2   Proofs from Section 4 (Sorting)

*Proof of Theorem 2.* In the proof of Theorem 2 in [14] lower bounds for probabilities of improving the current fitness have been established. For the function INV there are $m := \binom{n}{2}$ fitness levels. Using the straightforward partition $A_i := \{x \mid f(x) = i\}$, the probability of an improvement on fitness level $m - i$ is at least $3i/(2en(n-1))$. Applying the first result from Theorem 1 we get the following upper bound for the parallel expected time of an island model with an arbitrary topology. Using $\sum_{i=1}^{m-1} 1/i^{1/2} \leq \int_0^m 1/i^{1/2} \, \mathrm{d}i = 2m^{1/2}$,

$$\begin{aligned}
E(T^{\mathrm{par}}) &\leq 2 \sum_{i=0}^{m-1} \frac{1}{s_i^{1/2}} + \frac{1}{\mu} \sum_{i=0}^{m-1} \frac{1}{s_i} \\
&\leq \frac{2n}{\alpha^{1/2}} \sum_{i=1}^{m} \frac{1}{i^{1/2}} + \frac{n^2}{\alpha\mu} \sum_{i=1}^{m} \frac{1}{i} \\
&\leq \frac{4n}{\alpha^{1/2}} \cdot m^{1/2} + \frac{n^2}{\alpha\mu} \cdot H(m) \\
&= O\left(n^2 + \frac{n^2 \log n}{\mu}\right) .
\end{aligned}$$

For HAM, LAS, and EXC only fitness values in $\{0, \ldots, n\}$ are possible [14]. The probability for the (1+1) EA making an improvement is bounded from below by $s_i \geq 1/(en(n-1)) \geq 1/(en^2)$ for HAM and EXC and by $s_i \geq 1/(2en(n-1)) \geq 1/(2en^2)$. We thus get $s_i \geq \alpha/n^2$ in all cases when choosing $\alpha \in \{1/e, 1/(2e)\}$ appropriately. For ring graphs Theorem 1 results in the bound

$$\begin{aligned}
E(T^{\mathrm{par}}) &\leq 2 \sum_{i=1}^{n-1} \frac{1}{s_i^{1/2}} + \frac{1}{\mu} \sum_{i=1}^{n-1} \frac{1}{s_i} \\
&\leq \frac{2n}{\alpha^{1/2}} \sum_{i=1}^{n} \frac{1}{i^{1/2}} + \frac{n^2}{\alpha\mu} \sum_{i=1}^{n} \frac{1}{i} \\
&\leq \frac{4n}{\alpha^{1/2}} \cdot n^{1/2} + \frac{n^{3/2}}{\alpha\mu} \cdot H(n) \\
&= O\left(n^2 + \frac{n^2 \log n}{\mu}\right) .
\end{aligned}$$

For torus or grid graphs we get, using $\sum_{i=1}^{m-1} 1/i^{1/3} \leq \int_0^m 1/i^{1/3} \, \mathrm{d}i = 1.5 \cdot m^{1/3}$,

$$\begin{aligned}
E(T^{\mathrm{par}}) &\leq 3 \sum_{i=1}^{n-1} \frac{1}{s_i^{1/3}} + \frac{1}{\mu} \sum_{i=1}^{n-1} \frac{1}{s_i} \\
&\leq \frac{3n}{\alpha^{1/3}} \sum_{i=1}^{n} \frac{1}{i^{1/3}} + \frac{n^2}{\alpha\mu} \sum_{i=1}^{n} \frac{1}{i} \\
&\leq \frac{4.5n}{\alpha^{1/3}} \cdot n^{1/3} + \frac{n^2}{\alpha\mu} \cdot H(n) \\
&= O\left(n^{4/3} + \frac{n^2 \log n}{\mu}\right) .
\end{aligned}$$

Finally, for $K_\mu$ the result is

$$E(T^{\mathrm{par}}) \leq n + \frac{1}{\mu} \sum_{i=1}^{n-1} \frac{1}{s_i} \leq n + \frac{n^2}{\alpha\mu} \sum_{i=1}^{n} \frac{1}{i} = O\left(n + \frac{n^2 \log n}{\mu}\right) . \qquad \square$$

In contrast to HAM, LAS, and EXC, the upper bound for INV used a large number of $\Theta(n^2)$ fitness levels. This had led to rather loose upper bounds for parallel optimization times. The informed reader might think that grouping fitness values to create fewer, larger fitness levels could yield better upper bounds for INV. A requirement for this to work is that mutations must still improve the fitness by so much that the current fitness level is left. However, this is not always possible for INV. Consider the permutation

$$\left(\frac{n}{2}+1,\ 1,\ \frac{n}{2}+2,\ 2,\ \frac{n}{2}+3,\ 3,\ \ldots,\ n,\ \frac{n}{2}\right).$$

The difference between its fitness and the fitness of the optimum is $\Theta(n^2)$. This large value suggests that large improvements are possible. But for the above permutation, every elementary operation increases the fitness by only $O(1)$. This indicates that parallelization does not always lead to drastic speedups for INV.

### A.3  Proofs from Section 5 (Shortest Paths)

*Proof of Theorem 3.* We say that a vertex is optimized in case a shortest path to this vertex has been found. Due to the fitness function, such a shortest path can never be lost. As we are dealing with a multiobjective formulation of the problem, we cannot directly apply the fitness level method. Instead, we use this method for estimating the time until layers of vertices have been optimized.

As in [14] we define $\ell_i$ as the maximum number of edges on any shortest path from the source $s$ to node $i$. We consider layers of vertices with the same $\ell$-value. Then $n_j = \#\{i \mid \ell_i = j\}$ describes the number of vertices on the $j$-th layer, i.e., vertices where all shortest paths have at most $j$ edges.

Once all layers $1, \ldots, j-1$ have been optimized, each vertex $v$ in Layer $j$ becomes optimized if a predecessor $w$ on a shortest path is found for $v$. This is because a shortest path from $s$ to $w$ plus the edge $(w, v)$ gives a shortest path to $v$. The probability of setting a correct predecessor for $v$ and not changing any other predecessor is at least $1/(en^2)$. If $i$ vertices in Layer $j$ are not optimized yet, the probability of increasing the number of optimized vertices is at least $s_i := i/(en^2)$. Applying the fitness-level method (Theorem 1) for each layer and noting there are at most $\ell := \max\{j \mid n_j > 0\}$ layers yields the following. For the ring graph or any other strongly connected topology

$$\begin{aligned}
\mathrm{E}\left(T^{\mathrm{par}}\right) &\leq \sum_{j=1}^{\ell} 2 \sum_{i=1}^{n_j} \frac{1}{s_i^{1/2}} + \sum_{j=1}^{\ell} \frac{1}{\mu} \sum_{i=1}^{n_j} \frac{1}{s_i} \\
&= 2e^{1/2} n \sum_{j=1}^{\ell} \sum_{i=1}^{n_j} \frac{1}{i^{1/2}} + \frac{en^2}{\mu} \sum_{j=1}^{\ell} \sum_{j=1}^{n_j} \frac{1}{i} \\
&\leq 4e^{1/2} n \sum_{j=1}^{\ell} n_j^{1/2} + \frac{en^2}{\mu} \sum_{j=1}^{\ell} \ln(en_j).
\end{aligned}$$

As $\sum_{j=1}^{\ell} n_j = n$ and both functions $\sqrt{x}$ and $\ln(x)$ are concave, the worst case for both $\sum$-terms is attained for $n_1 = \cdots = n_\ell = n/\ell$. This yields

$$\mathrm{E}\left(T^{\mathrm{par}}\right) \leq 4e^{1/2} n^{3/2} \ell^{1/2} + \frac{en^2 \ell \ln(en/\ell)}{\mu}.$$

For the torus we get, using $\sum_{i=1}^{m-1} 1/i^{1/3} \leq \int_0^m 1/i^{1/3}\, \mathrm{d}i = 1.5 \cdot m^{1/3}$,

$$\mathrm{E}\left(T^{\mathrm{par}}\right) \leq \sum_{j=1}^{\ell} 3 \sum_{i=1}^{n_j} \frac{1}{s_i^{1/3}} + \sum_{j=1}^{\ell} \frac{1}{\mu} \sum_{i=1}^{n_j} \frac{1}{s_i}$$

$$= 3 e^{1/3} n^{2/3} \sum_{j=1}^{\ell} \sum_{i=1}^{n_j} \frac{1}{i^{1/3}} + \frac{en^2}{\mu} \sum_{j=1}^{\ell} \sum_{j=1}^{n_j} \frac{1}{i}$$

$$\leq 4.5 e^{1/2} n^{2/3} \sum_{j=1}^{\ell} n_j^{2/3} + \frac{en^2}{\mu} \sum_{j=1}^{\ell} (\ln(n_j) + 1)$$

$$\leq 4.5 e^{1/2} n^{4/3} \cdot \ell^{1/3} + \frac{en^2 \ell \ln(en/\ell)}{\mu}.$$

For the complete graph $K_\mu$ we get

$$\mathrm{E}\left(T^{\mathrm{par}}\right) \leq \sum_{j=1}^{\ell} n_j + \sum_{j=1}^{\ell} \frac{1}{\mu} \sum_{i=1}^{n_j} \frac{1}{s_i} \leq n + \frac{en^2 \ell \ln(en/\ell)}{\mu}.$$

□

### A.4 Proofs from Section 6 (Eulerian Cycles)

We first prove Theorem 5 and then reuse some of the proof arguments for proving Theorem 4.

*Proof of Theorem 5.* For every island, unless the current walk has formed a cycle, there is always at least one jump operation that increases the length of the walk. The probability of such a jump is at least $1/m^2$. By Chernoff bounds with probability $1 - e^{-\Omega(m)}$ after $\tau \geq m^3$ generations an island either once has reached a walk that forms a cycle of length $m/2$ or its current walk is strictly longer than $m/2$.

Assume that the above condition holds for all islands. We estimate the probability that all islands have reached a cycle of $m/2$ edges. Due to independence, we can focus on RLS running on a single island.

One important observation for RLS is that once RLS has discovered a walk $(e_1, \ldots, e_\ell)$ of length $\ell \geq 2$ the current walk will always contain the edges $e_1, \ldots, e_\ell$. So, after the first walk of length at least 2 is discovered, it will typically be extended by prepending a matching edge $e_0$ or appending a matching edge $e_{\ell+1}$. (In the latter case the walk can grow by more than one edge in case the sequence of edges happens to continue with a proper edge $e_{\ell+2}$, and so on.) Once the current walk contains edges from both $C$ and $C'$, the global optimum can be found easily as there is always at least one jump operation extending the current walk. The expected remaining time until a Eulerian cycle is constructed is bounded by $m^3$.

Assume pessimistically that the first walk of length at least 2 lies completely in $C$ (say). Consider the first point of time where $v^*$ becomes part of the walk. Suppose that the walk starts with $v^*$ and that the walk is not equal to $C$. Then there are three edges that can jump to the first position of the current walk: the edge in $C$ incident to the first edge of the walk and two edges in $C'$ that contain $v^*$. As all jumps are equally likely, the probability that a jump adds one of the edges in $C'$ before the edge from $C$ is added is $2/3$. Symmetric arguments apply if the walk ends with $v^*$. There is one caveat, though. The jump that has added the edge of the current walk leading to $v^*$ can have added further edges. This would mean that the current walk has already extended past $v^*$, depending on the edges following in the edge sequence. However, all three mentioned edges so far have been symmetric to the algorithm in a sense that their order has not had an impact on the fitness so far. Therefore, each of these is equally likely to be in the position of the next edge in the edge sequence. So we again have a probability of $2/3$ that the walk has been extended towards $C'$.

The probability that at least one island makes the right decision is $1 - 3^{-\mu}$. If this happens, the expected remaining optimization time is $O(m^3)$ as shown above. If this does not happen, we resort to the general upper bound $O(m^4)$ by Neumann [11] for the time until an improvement is found. This proves the claimed bound $O(m^3 + 3^{-\mu} \cdot m^4)$. □

In order to prove Theorem 4 we first need the following lemma about the concentration of hitting times for fair random walks on integers. It follows using standard Chernoff bounds.

**Lemma 1.** *For the fair random walk on $\mathbb{Z}$, starting in state 0, define $T(k)$, $k \in \mathbb{N}$, as the first hitting time of a state in $\{-k, +k\}$. We have $Pr(T(k) = t) \leq 2e^{-k^2/t}$ if $t > 2k$ and $Pr(T(k) = t) \leq 2(e/4)^k$ if $t \leq 2k$.*

*Consequently, $Pr(T(k) \leq t) \leq 2t(e/4)^k$ if $t \leq 2k$ and*
$Pr(T(k) \leq t) \leq 4k(e/4)^k + 2te^{-k^2/t}$ *if $t > 2k$.*

*Proof.* As the claimed bounds for $Pr(T(k) = t)$ are non-decreasing with $t$, the second statement follows from the first one and the union bound.

The proof of the first statement is a simple application of Chernoff bounds. Let $X$ be the random number of steps among the first $t$ iterations of the random walk where the current state is increased. Clearly, $E(X) = t/2$ and one of the two target states is reached in $T$ steps if and only if $X = t/2 + k$ or $X = t/2 - k$. The probabilities for the last two events are equal, hence $Pr(T(k) = t) = 2Pr(X = t/2 + k)$ and we only need to estimate the last probability.

We have $t/2 + k = (1 + \delta) \cdot \mathrm{E}(X)$ for $\delta = 2k/t$. If $t > 2k$ we use a well-known Chernoff bound for $0 < \delta < 1$ and have

$$\Pr(X = t/2 + k) \leq \Pr(X \geq t/2 + k)$$
$$\leq e^{-t/2 \cdot (2k/t)^2/2} = e^{-k^2/t}.$$

For $t \leq 2k$ we have

$$\Pr(X = t/2 + k) \leq \Pr(X \geq t/2 + k)$$
$$\leq \left(\frac{e^\delta}{(1+\delta)^{1+\delta}}\right)^{\mathrm{E}(X)}$$
$$= \left(\frac{e^{2k/t}}{(1+2k/t)^{1+2k/t}}\right)^{t/2}$$
$$= \left(\frac{e}{(1+2k/t)^{t/(2k)+1}}\right)^k \leq (e/4)^k$$

as $2k/t \geq 1$ and the function $(1+x)^{1/x+1}$ is monotonically increasing for $x \geq 1$. □

*Proof of Theorem 4.* The proof consists of two parts. We first prove that with constant probability the island model will reach a state where all islands need to rotate a cycle of length $m/2$ by a distance of $\Theta(m)$. As all islands have the same fitness, we can safely ignore migration until the first island has found an improvement after rotating the cycle. The time it takes to get there will establish the lower bound.

Consider the first point of time $t^*$ where an island extends its walk past $v^*$. We know by the proof of Theorem 5 that there is a chance of $1/3$ that the walk will continue in the same cycle. The probability that during the next $\tau \cdot \mathrm{diam}(T)$ generations following $t^*$ no island makes a further improvement—and no other island makes a simultaneous improvement at time $t^*$—is at least

$$\left(1 - \frac{6}{m(m-1)}\right)^{(\tau \operatorname{diam}(T)+1)\mu} \geq \Omega(1)$$

since there are always at most 6 improving jump operations and $\tau \operatorname{diam}(T) \mu = O(m^2)$. After this time all islands will have been taken over by the same solution. Hence, we have a probability of $\Omega(1)$ that one island extends its walk past $v^*$, stays in the same cycle, and communicates this superior solution to all other islands. This means that the present edges of the walk will be maintained on each island until the first island has closed a cycle of $m/2$ edges.

By the same argument, with again probability $\Omega(1)$ and independently from the previous events, we have that the island that first closes this cycle is the only island where an improvement happens. Again this solution takes over all islands. Now we use the following argument on symmetry. So far all islands have behaved as if the instance consisted only of $C$. Due to the perfect symmetry of the subgraph induced by $C$, each vertex is equally likely to be the start and end vertex of a walk covering $C$. With probability at least $1/2$, again independently from previous events, we have that this vertex has distance at least $m/8$ from $v^*$ on the island that takes over the system. Then a rotation of the walk by a distance of $\Theta(m)$ is necessary for a further improvement.

Following Neumann [11], the only accepted operations for all islands are rotations of the cycle, unless it starts and ends in $v^*$. As no direct fitness improvements are possible and only strictly better immigrants are accepted, all islands evolve independently until an island finds an improvement.

A step rotating the current cycle is called a relevant step. It has probability $2/(m(m-1))$ and the probability of having more than $t := bm^2/(\ln \mu)$ relevant steps, for some constant $b > 0$ specified later, in $bm^3(m-1)/(3 \ln \mu)$ generations is $e^{-\Omega(m)}$ by Chernoff bounds. Assume in the following that each island makes at most $t$ relevant steps, which happens with probability at least $1 - \mu \cdot e^{-\Omega(m)}$.

A clockwise rotation has the same probability as a counterclockwise rotation. If we map the possible positions of the start/end of the cycle to $\mathbb{Z}$ such that after takeover each island starts at 0, each island performs a fair random walk. This random walk has to cover a distance of at least $m/8$ from 0 in order to reach $v^*$. We apply Lemma 1 with $k := m/8$. The probability of reaching this goal in $t := bm^2/(\ln \mu)$ steps, $b > 0$ an appropriate constant, is at most $1/(2\mu)$. By the union bound, the probability that any island has reached this goal is at most $1/2$. So, with probability at least $1/2 - \mu \cdot e^{-\Omega(m)}$ the island model has not found an improvement after $bm^3(m-1)/(3 \ln \mu) = \Omega(m^4/(\log \mu))$ generations. As $1/2 - \mu \cdot e^{-\Omega(m)} = \Omega(1)$, this establishes the claimed lower bound. □

Using symmetrically restricted jumps decreases the number of possible jump operations from $m(m-1)$ to only $2m$. Recall that for revolving a cycle, only two jump operations are possible. These jumps are still feasible with the restricted operator and they now have a higher probability of $1/(2m)$ each. This raises the probability of making a relevant step to $1/m$. By exactly the same reasoning as above, only changing the period of $bm^3(m-1)/(3 \ln \mu)$ generations towards $bm^3/(3 \ln \mu)$ generations, we arrive at the results shown in the second line of Table 3.